\documentclass[letterpaper, 10 pt, conference]{ieeeconf}  

\IEEEoverridecommandlockouts                              

\usepackage{graphics} 
\usepackage{epsfig} 
\usepackage{mathptmx} 
\usepackage{times} 
\usepackage{amsmath} 
\usepackage{amssymb}  
\usepackage{subcaption}
\usepackage{algpseudocode}
\usepackage{algorithmicx}
\usepackage{algorithm}
\title{\LARGE \bf
Semantic Zone based 3D Map Management for Mobile Robot
}

\author{Huichang Yun$^{1}$ and Seungho Yoo$^{1*}$%
\thanks{$^{1}$ Dept. of computer Engineering, Pukyong National University}%
\thanks{$^{*}$ Corresponding Author}%
}

\begin{document}

\maketitle
\thispagestyle{empty}
\pagestyle{empty}

\begin{abstract}

 Mobile robots operating in large-scale indoor environments such as hospitals and logistics centers require accurate 3D spatial representation to perceive complex structures. However, 3D maps incur substantial memory consumption, making it challenging to maintain complete map data within the robot's limited computational resources. Although existing SLAM frameworks implement memory management techniques such as graph pruning, keyframe culling, and working memory (WM) constraints—these approaches typically rely on geometric distance or temporal metrics for memory retrieval decisions, often resulting in inefficient data loading patterns in spatially compartmentalized environments.

To address this fundamental limitation, this thesis proposes a semantic zone-based 3D Map Management method that fundamentally shifts the memory management paradigm from geometry-centric to semantics-centric control. Rather than managing map components based on geometric proximity alone, the proposed approach partitions the environment into semantically meaningful spatial units—such as lobbies, hallways, and patient rooms—and designates these zones as the primary unit of memory management. Specifically, the method establishes a systematic mapping between map signatures/keyframes and their respective semantic zones, enabling the system to dynamically load only task-relevant zones into working memory (WM) based on the robot's current location and operational requirements, while systematically unloading inactive zones to Long-Term Memory (LTM). This semantic zone based control strategy enables the system to strictly enforce user-defined memory thresholds MemoryThr.

The proposed method was implemented by integrating a zone-aware memory retrieval and unload algorithm into the RTAB-Map framework, maintaining compatibility with the existing WM/LTM hierarchical architecture. Quantitative evaluation against standard RTAB-Map memory management reveals that the semantic zone-based approach substantially reduces unnecessary signature load/unload cycles and cumulative memory utilization. The results demonstrate that semantic zone-based management maintains stable, predictable memory usage while preserving map availability for critical navigation and localization tasks. Our code is open-sourced at: https://github.com/huichangs/rtabmap/tree/segment

\end{abstract}

\section{INTRODUCTION}

For autonomous robots to perform tasks in large-scale indoor environments such as hospitals, warehouses, and factories, a detailed understanding of the surrounding space is essential. Robots employ maps to represent their environment, which can be classified as 2D or 3D representations. While 2D occupancy grid maps are useful for path planning on flat surfaces, they have inherent limitations in representing three-dimensional terrain and obstacles such as ramps, stairs, and low obstacles. Due to these limitations, 3D mapping technology is becoming increasingly important in practical service robot applications for directly representing three-dimensional environmental structures.

The primary challenge with 3D maps is that while they contain significantly richer information than 2D maps, their memory consumption is substantially higher. When representing large indoor environments composed of corridors, rooms, lobbies, and storage areas using 3D maps, maintaining all 3D map data in memory becomes a constraint for mobile robots with limited hardware resources.

Existing SLAM frameworks have implemented various memory management techniques to mitigate this issue. The ORB-SLAM family of SLAM systems periodically removes redundant or low-contribution keyframes and map points to limit graph size, while graph-based SLAM research has proposed methods such as information-theoretic graph pruning or submodular optimization-based keyframe selection. RTAB-Map advances this approach by introducing a hierarchical STM-WM-LTM structure and a sophisticated memory management system that loads and unloads signatures between WM and LTM. This enables long-term mapping by maintaining the complete map in a database while keeping only necessary signatures in working memory for real-time processing.

Although numerous research efforts have focused on submap or region-based approaches, hierarchical SLAM for improved loop closure performance, and semantic information utilization, no prior work has leveraged semantic information as a memory management mechanism to control which map components remain in memory.

This thesis proposes a semantic zone-based 3D Map Management technique that is not dependent on any specific SLAM algorithm and can be applied to existing 3D mapping systems including ORB-SLAM, RTAB-Map, and Nvblox. Specifically, targeting indoor hospital environments, the proposed approach partitions the environment into meaningful spatial units such as lobbies, corridors, and patient rooms, establishes a mapping between signatures and semantic zones, and performs memory management at the zone level.

The implementation utilizes RTAB-Map and evaluates performance by comparing the accumulated signature load/unload operations and cumulative memory usage against the baseline RTAB-Map retrieval algorithm. The primary contributions of this research are:

\begin{enumerate}
\item \textbf{Semantic Zone-Based 3D Map Memory Management Concept:} This work partitions dense 3D maps into semantic zones representing meaningful spaces such as corridors and rooms, and uses zones as the primary memory management unit. Unlike existing keyframe or voxel-centric approaches, this provides a new perspective by controlling memory based on semantically meaningful spatial regions.

\item \textbf{Integration with RTAB-Map Memory Framework:} The proposed technique modifies only the retrieve and remove algorithms while maintaining the existing WM/LTM structure and SLAM mechanisms of RTAB-Map. This demonstrates that the proposed approach can be applied at the policy level without requiring fundamental changes to existing SLAM frameworks.

\item \textbf{Hierarchical 3D Mapping and Hospital Scenario-Based Evaluation:} Using Isaac Sim's Amazon Hospital World, this work implements a hierarchical 3D mapping strategy where common areas like lobbies and corridors are mapped first, followed by individual patient rooms added as layers. Experimental comparison of standard RTAB-Map and the proposed semantic zone-based approach under identical memory thresholds quantitatively demonstrates reduction in unnecessary signature load/unload operations.

\end{enumerate}

The remainder of this paper is organized as follows. Section II provides background on 3D mapping and SLAM systems. Section III reviews related work on 3D map management and semantic information utilization. Section IV presents the design and algorithms of the proposed semantic zone-based memory management technique. Section V describes the implementation and evaluation scenarios. Section VI presents experimental results, Section VII discusses limitations and future work, and Section VIII concludes the paper.

\section{BACKGROUND}

For robots to perform tasks in indoor environments, understanding three-dimensional space is essential. While 2D occupancy grid maps are useful for planar navigation, they are limited in environments requiring three-dimensional information such as ramps, stairs, and low obstacles. Therefore, recent research has increasingly focused on 3D map generation that can represent real environmental information. This section introduces 3D mapping and representative SLAM frameworks relevant to this research.

\subsection{3D Mapping and SLAM Systems}

This section briefly introduces representative SLAM and 3D mapping frameworks directly related to this research. Specifically, we focus on ORB-SLAM as a representative feature-point based SLAM, RTAB-Map which implements graph-based SLAM with hierarchical memory structures, and nvblox, a GPU-accelerated dense 3D reconstruction library, detailing the types of maps each framework generates and their relevance to this research.

\subsubsection{ORB-SLAM (Oriented FAST and Rotated BRIEF SLAM)}

ORB-SLAM is the most representative framework for feature-based Visual SLAM, proposed by Mur-Artal et al. It extracts ORB (Oriented FAST and Rotated BRIEF) features from images and performs matching, supporting monocular, stereo, and RGB-D cameras. The most distinctive characteristic of this system is the generation of sparse maps. By focusing primarily on feature points such as corners and object edges, the system uses minimal memory and achieves fast computation with excellent real-time performance. With evolution to ORB-SLAM3, it now supports VIO (Visual-Inertial Odometry) with IMU sensors and multi-session map merging capabilities, significantly improving robustness.

However, sparse maps contain only 3D feature points and do not represent textured regions such as walls and floors. In path planning tasks, it is difficult to accurately represent free space and obstacles, making the sparse map alone insufficient for direct collision avoidance. In practice, robots use additional occupancy grid maps or voxel maps for path planning rather than relying solely on ORB-SLAM's sparse map.

From a memory management perspective, while ORB-SLAM does not employ a hierarchical WM/LTM structure like RTAB-Map, it controls map capacity through internal culling strategies for keyframes and map points. For example, it periodically removes redundant or low-observation-value keyframes and deletes unused map points, preventing the graph from growing indefinitely. This approach manages memory by keeping the graph itself small rather than managing components hierarchically.

\subsubsection{RTAB-Map (Real-Time Appearance-Based Mapping)}

RTAB-Map is a framework combining appearance-based loop closure detection with graph-based SLAM structures, supporting various sensors including RGB-D, stereo, monocular cameras, and LiDAR, and can generate various map formats including 2D grid maps, 3D point clouds, and meshes.

The basic unit of RTAB-Map is the signature, with each signature containing sensor data (images, depth, laser), pose estimates, feature descriptors, and connection information with adjacent nodes. Signatures are added as nodes to the pose graph, and results from odometry, loop closure, and scan matching are represented as edges. The graph optimizer performs nonlinear optimization whenever new constraints arrive to compute globally consistent trajectories and maps.

The core innovation of this system is its memory management technique inspired by human memory mechanisms. RTAB-Map distinguishes between overall map storage in working memory (WM) and Long-Term Memory (LTM). Recent data and frequently observed regions required for loop closure detection and localization are maintained in WM to ensure real-time processing, while older or infrequently used data are transferred to LTM (database) to reduce memory burden. This addresses the fundamental SLAM problem of infinite data accumulation over time and enables long-term mapping within limited resources.

\subsubsection{Nvblox (GPU-Accelerated Distance Fields)}

Nvblox is a recently developed mapping framework by NVIDIA that actively leverages GPU acceleration to generate high-resolution dense maps in real-time. Unlike traditional CPU-based mapping approaches limited by computational capacity and forced to use low-resolution voxel maps, Nvblox constructs TSDF (Truncated Signed Distance Field) and ESDF (Euclidean Signed Distance Field) on the GPU to build high-resolution 3D maps. However, such high-resolution dense maps consume substantial GPU memory, necessitating memory management strategies similar to those discussed for nvblox.

Nvblox functions less as an independent SLAM algorithm and more as a mapping module that constructs high-quality 3D maps based on pose information provided externally. Recently, it has been increasingly used in combination with robot operating systems like Isaac ROS to maximize perception capabilities of autonomous robots. Nvblox's memory management is currently performed through simple distance-based activation/deactivation strategies, representing regions densely when near the robot and sparsely when beyond certain thresholds, or deactivating blocks for memory management.

\section{RELATED WORK}

This section reviews research on 3D map and submap-based approaches as well as semantic-based methods to understand current trends in addressing 3D map limitations.

\subsection{3D Map and Submap-Based Approaches}

To achieve scalability in large-scale SLAM rather than using single monolithic maps, active research has explored using multiple maps or submaps. Aguiar et al. proposed memory optimization techniques for localization and mapping based on topological maps, representing the environment as a topological map with nodes and edges, then activating specific node sets for localization while keeping others inactive to reduce memory and computation. This approach is conceptually similar to the zone concept in this research in that it manages maps at a node level, but focuses more on determining which topological nodes to use for current localization rather than the actual load/unload mechanisms of map data.

Ehlers et al. proposed a Map Management Approach for large-scale indoor and outdoor SLAM environments, using submap-level structure and state management, distinguishing between active and inactive submaps to maintain only necessary submaps in memory for scalability. Similar approaches by Taranta et al. and recent Region-based SLAM-aware exploration techniques divide environments into regions or areas and control SLAM operations and exploration strategies by region.

These research efforts share the common principle of dividing space into multiple submaps, regions, or areas and selectively activating portions as needed, but they do not address semantic information-based keyframe-level load/unload memory management mechanisms as this thesis proposes.

From a 3D dense map perspective, multi-resolution and submap-based maps have been key research directions for memory efficiency and scalability. Stückler and Behnke proposed multi-resolution surfel maps with layers of different resolutions enabling high-resolution representation for regions of interest. Tang et al. proposed multi-implicit-submaps structures for scalable and robust online RGB-D reconstruction, while Tian et al. achieved memory-efficient neural implicit reconstruction through multiresolution submap optimization. HRGS introduced hierarchical Gaussian splatting for memory-efficient high-resolution 3D reconstruction.

These approaches reduce memory usage by designing the map representation itself using multi-resolution, multiple implicit submaps, or hierarchical Gaussians, differing from this paper's approach of controlling which parts to keep in memory at the policy level above existing SLAM framework WM/LTM hierarchies.

Graph-based SLAM memory management has also been extensively researched in directions of reducing graph and keyframe numbers. Kretzschmar et al. proposed information-theoretic graph pruning to sparsify graphs by removing low-information-contribution edges. Thorne et al. researched keyframe selection and usage strategies using submodular optimization. The MS-SLAM system implements memory-efficient visual SLAM through sliding window map sparsification, while Venkatesh studied automatic memory management for ORB-SLAM3, automating keyframe and map point creation/deletion policies to mitigate memory issues in long-term operations.

RTAB-Map introduced explicit keyframe load/unload mechanisms between memory and database hierarchies with its WM/LTM structure. Subsequent RTAB-Map research has been among the few works directly addressing keyframe-level memory management.

\subsection{Semantic Information with 3D Maps}

Recent research increasingly incorporates semantic information into map representations and localization. SegMap extracts segments from 3D point clouds and learns data-driven descriptors for each segment, enabling segment-level compressed map representation and place recognition. SemSegMap extends this by integrating color and semantic labels into descriptors for segment-based semantic localization. RoboHop constructs segment-based topological maps and leverages segment-level representations for open-world visual navigation.

Hofmann et al. proposed frameworks for efficient semantic mapping in dynamic environments, adjusting map update and maintenance strategies based on environmental semantic structures. Mora et al. proposed techniques using Voronoi diagrams to analyze free and occupied space structures and partition maps into meaningful regions. These regional partitioning methods relate to the semantic zone concept in this paper by defining meaningful map regions based on semantic or geometric structures.

Scucchia and Maltoni proposed region prediction techniques for efficient large-scale localization. The Region-based SLAM-aware exploration research similarly divides environments into multiple regions, designing exploration strategies by region, combining region-based (topological/region-based) representations with SLAM and exploration. However, the focus lies on determining which region to perform localization and exploration in, not extending to semantic control of memory load/unload policies.

Zheng et al. proposed semantic SLAM systems utilizing large visual models, combining semantic perception and map construction in complex environments. S-Graphs 2.0 proposes hierarchical and semantic graph structures, performing SLAM optimization and loop closure integrating geometric and semantic information.

These semantic mapping and semantic SLAM research efforts focus on (1) segment/object-level semantic map representations and robust loop closure, (2) semantic graph-based topological representations and localization, (3) combining large model-based high-precision semantic recognition. However, in most cases, semantic information serves to improve map representation quality and perception/navigation quality, unlike this paper which uses semantic information as the basis for determining which zones to maintain in memory.

\section{SEMANTIC ZONE BASED 3D MAP MANAGEMENT}

This section explains the semantic zone-based 3D map management technique proposed in this research. The core innovation is not maintaining an entire large 3D dense map in memory, but instead partitioning the large indoor environment into semantic zones and implementing a new memory management approach that loads and unloads map data at the zone level.

This memory management technique is not dependent on specific SLAM algorithms or 3D reconstruction methods. The zone concept can be composed of groupings of ORB-SLAM keyframes, RTAB-Map signature IDs, nvblox blocks, or similar structures, and can be applied as an upper layer across various framework map representations.

\subsection{Overview}
\begin{figure}[t]
    \centering
    \centerline{\includegraphics[width=80mm]{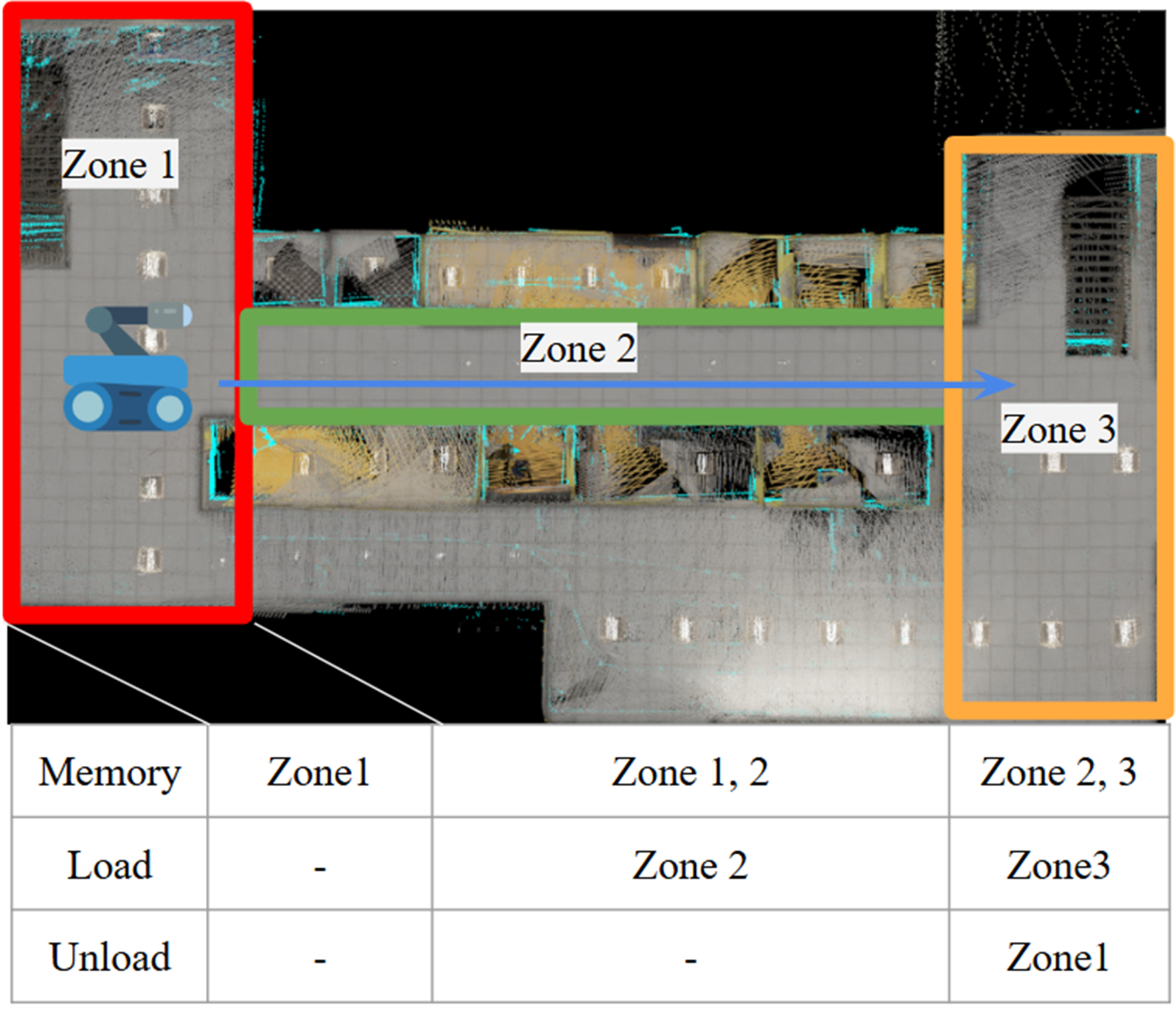}}
    \caption{Overview of semantic zone based 3d map management approach}
    \label{figure_1} 
\end{figure}

Fig. 1. conceptually illustrates how the zone composition in memory changes as the robot traverses multiple zones while respecting set memory thresholds through zone load/unload operations.

Initially, the robot operates only within Zone 1, with 3D map data corresponding to Zone 1 loaded in memory. As the robot moves toward the boundary between Zone 1 and Zone 2, Zone 2 information is required for future movement, so Zone 2 is loaded. At this point, both Zone 1 and Zone 2 exist simultaneously in memory.

As the robot's path requires entry into Zone 3, it must load Zone 3 data. However, if the memory threshold prevents loading Zone 3, the proposed technique unloads Zone 1, which is unlikely to be used further, to free available memory and load Zone 3. Consequently, only Zone 2 and Zone 3 remain in memory.

In this manner, the proposed technique dynamically adjusts the set of active zones based on robot movement and task context, controlling memory usage by loading and unloading 3D map data based on this active zone set.

\subsection{Semantic Zone Definition}

Zone partitioning can be implemented through (1) geometric partitioning or (2) semantic partitioning based on spatial semantics. While this research targets semantic zone-based management, the advantages are illustrated by first examining the limitations of geometric partitioning.

\subsubsection{Geometric Partitioning Approach}
\begin{figure}[t]
    \centering
    \centerline{\includegraphics[width=80mm]{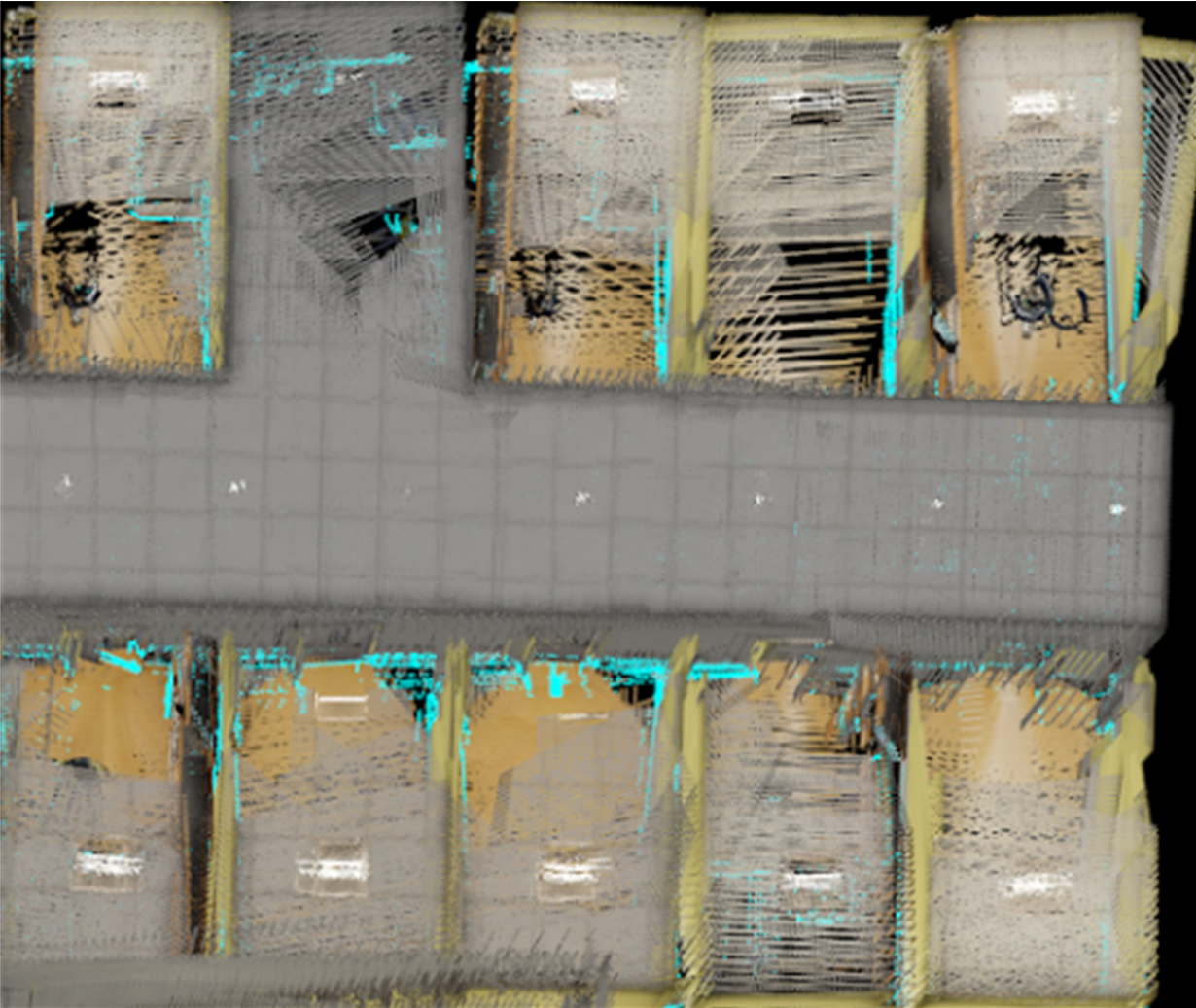}}
    \caption{Complex environment with many romms connected to the corridor}
    \label{figure_2} 
\end{figure}

Geometric partitioning divides map data based on physical distance or coordinate systems, with division by uniform voxel grids or blocks as typical examples. Alternatively, submaps are created when robot travel distance exceeds thresholds, or submaps are terminated when pose graph node counts reach predetermined sizes.

This approach is simple and easy to implement but has fundamental limitations in not considering actual environmental structure. For example, consider a robot navigating along a corridor in a complex space with many rooms connected to the corridor (Fig. 2). When dividing geometrically by fixed distance or grid size, multiple rooms adjacent to the corridor may be included in the same partitioning unit. In this case, even though the robot only navigates the corridor, it must load keyframes and dense map data from room interiors it never enters.

Such unnecessary loading increases memory usage and triggers additional unload operations to maintain memory thresholds, degrading overall memory management efficiency. As environment scale increases and room density grows, this phenomenon where geometrically unrelated spaces are unnecessarily loaded together becomes increasingly severe.

\subsubsection{Semantic Partitioning Approach}

Semantic partitioning defines zones based on spatial and functional semantics of environmental elements. For hospital maps, for example, naturally distinguished spaces from a human perspective include corridors, examination rooms, patient rooms, and lobbies. When zones are defined based on these semantic boundaries, each zone represents a single functional space with characteristic usage patterns.

The greatest advantage of semantic zone-based operation is the ability to load only actually needed spaces into memory. For example, when a robot performs corridor patrol tasks, only the corridor zone is activated while unvisited patient room zones remain unloaded, reducing unnecessary keyframe loading and maintaining memory thresholds more stably. Due to these advantages, this research employs semantic partitioning rather than geometric partitioning for zone-based map management.

\subsection{Semantic Zone-Based Memory Management}

Based on semantic zones defined in the previous section, this section explains the memory management technique for loading and unloading 3D maps at the zone level. The goal is to dynamically manage the set of zones that should be activated based on the robot's current tasks and location, control overall memory usage by maintaining only data from active zones in memory, while ensuring availability of necessary map data for navigation and localization tasks. The proposed memory management technique comprises three stages:

\subsubsection{Zone-Keyframe Mapping}

During map generation, record which zone each keyframe was acquired in, or identify all keyframes belonging to each zone from the complete set of keyframes constituting the map. For each zone, construct the set \(K_z\) of keyframes belonging to that zone. Expressed mathematically:

\[K_z = \{k \in K : k \in z\}\]

where \(K\) is the complete keyframe set and \(K_z \subseteq K\).

\subsubsection{Active Zone Set Determination}

Using Loop Closure Detection, externally obtained pose information, or similar methods, determine the set of zones \(Z_{active}\) that should be activated at the current moment. For example, when the robot is within Zone 1, \(Z_{active} = \{\text{Zone 1}\}\), and when approaching the boundary between Zone 1 and Zone 2, \(Z_{active} = \{\text{Zone 1, Zone 2}\}\).

\subsubsection{Zone-Based Load/Unload Policy Application}

Once \(Z_{active}\) is determined, check if it exceeds the memory threshold. With the system's total memory threshold as \(M_{max}\), ensure that the total memory usage of all keyframe sets belonging to active zones does not exceed \(M_{max}\). If exceeding this threshold appears likely, iteratively deactivate the least recently used zone from \(Z_{active}\) and unload its keyframes. Repeat this process until the memory usage of \(Z_{active}\) is below \(M_{max}\).

Through this semantic zone-based memory management, robots can maintain in memory only portions of zones necessary for current task and location, independent of overall environment scale.

\section{IMPLEMENTATION}

This section describes the simulation environment constructed to validate the proposed semantic zone-based memory management technique and the specific implementation of this approach on RTAB-Map. A virtual hospital environment was constructed based on NVIDIA Isaac Sim, with robots generating three-dimensional dense maps while traversing the environment. Subsequently, RTAB-Map's memory management module was extended to apply the semantic zone-based memory management technique described in Section IV.

\subsection{Isaac Sim Simulation Environment Setup}
\subsubsection{Geometric Partitioning Approach}

\begin{figure}[t]
    \centering
    \centerline{\includegraphics[width=80mm]{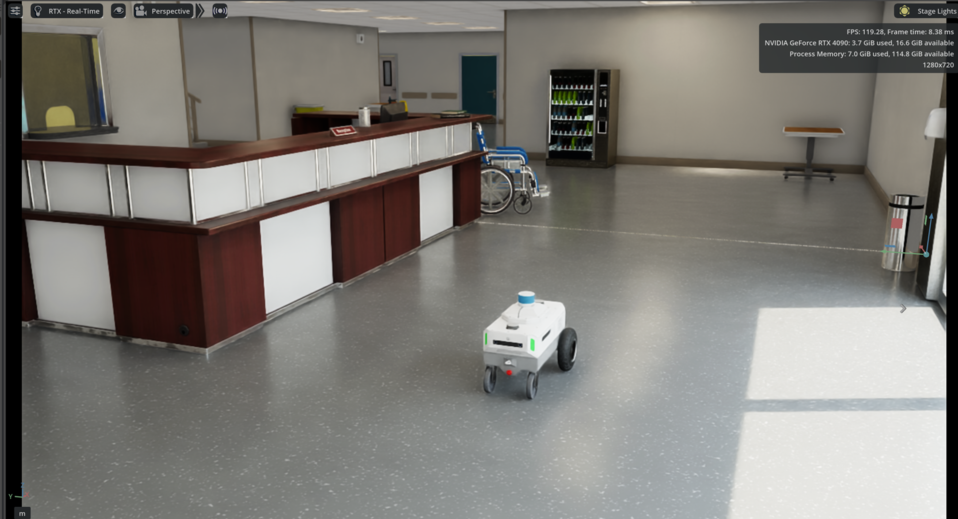}}
    \caption{Simulation Envirionment (Amazon Hospital World, Nova Carter)}
    \label{figure_2} 
\end{figure}

The simulation environment is shown in Fig. 3. It was constructed using the Amazon Hospital World provided by NVIDIA Isaac Sim, with the Nova Carter robot model equipped with RGB-D cameras and 2D LiDAR at the front.

The Amazon Hospital World has structure similar to actual hospitals and comprises various functional spaces including corridors, multiple rooms, and lobbies. This structure is suitable for evaluating the proposed semantic zone-based memory management technique. Because corridors and multiple rooms are arranged repeatedly, unnecessary room interior data frequently loads using purely geometric partitioning, while the effects of semantic zone partitioning are clearly observable.

\subsection{RTAB-Map Based Semantic Zone Memory Management Implementation}

RTAB-Map was selected as the foundational SLAM framework for implementation and validation of the proposed semantic zone-based memory management technique. RTAB-Map was chosen not simply because it is a widely known SLAM library, but because its memory management structure and modular design particularly suit the proposed technique's characteristics.

RTAB-Map provides an explicit memory structure distinguishing signatures between working memory (WM) and Long-Term Memory (LTM), maintaining signatures at frequencies below thresholds by unloading low-frequency or long-observed signatures to LTM and reloading them from the database when needed. This load/unload mechanism is already established at the framework level.

Furthermore, RTAB-Map remains the only work practically performing partial keyframe-based memory management between memory and database. While the recently released Nvblox includes memory management features, these are simple distance-based activation/deactivation strategies without WM/LTM structures and load/unload algorithms like RTAB-Map.

Therefore, performance evaluation primarily compares the standard RTAB-Map retrieve algorithm with the proposed semantic zone-based memory management technique under identical scenarios and memory thresholds, using signature load/unload frequency and accumulated loaded signature quantity as key metrics in Section VI to demonstrate how semantic zone-based management dramatically reduces unnecessary signature loading and unloading.

\subsection{Hierarchical 3D Mapping}

\begin{figure}[!t]
     \centering
     \begin{subfigure}[b]{0.40\textwidth}
        \centering
         \includegraphics[width=\linewidth]{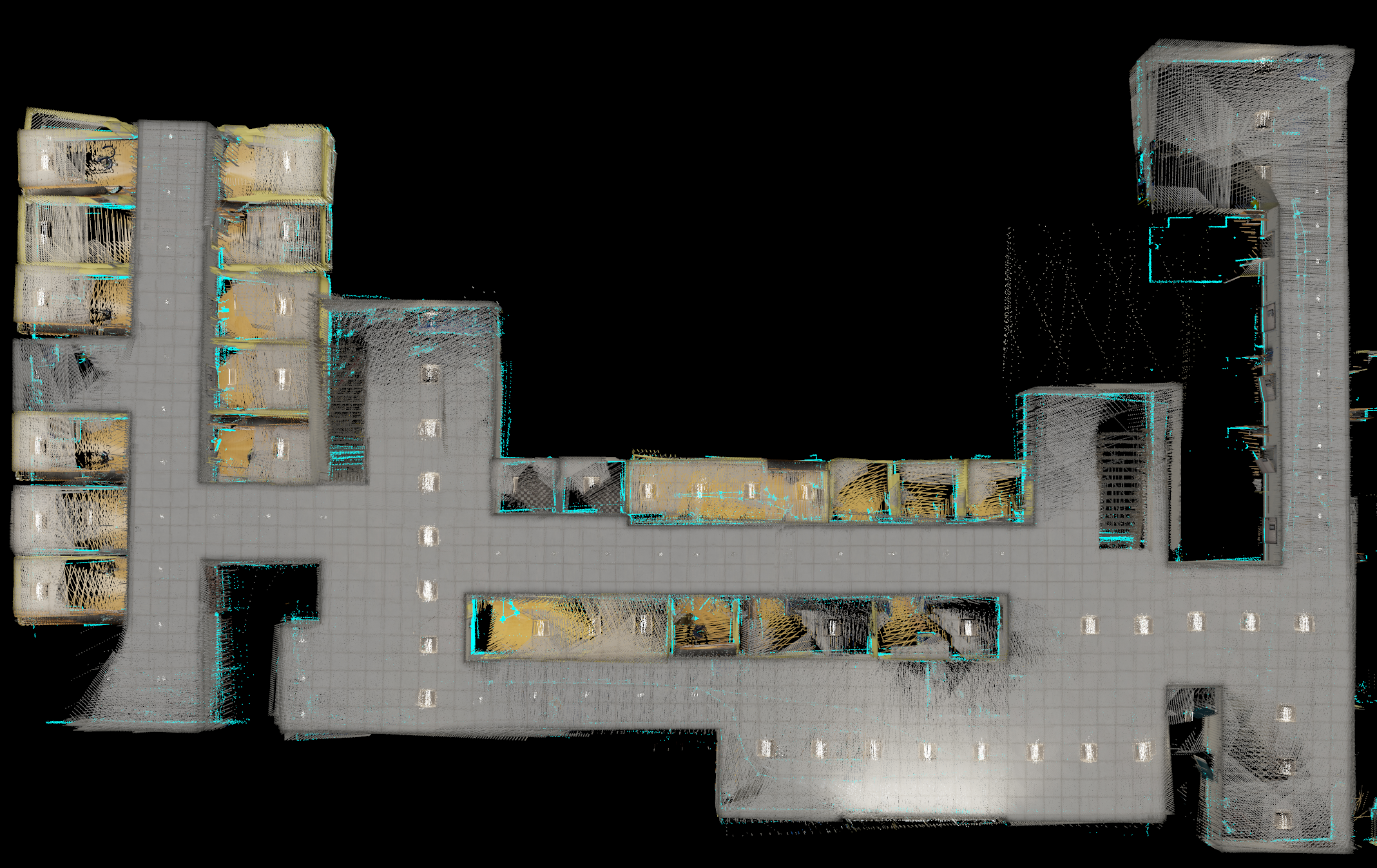}
         \caption{Amazon hospital world 3D map}
         \label{figure_4_a}
     \end{subfigure}
     \begin{subfigure}[b]{0.40\textwidth}
         \centering
         \includegraphics[width=\linewidth]{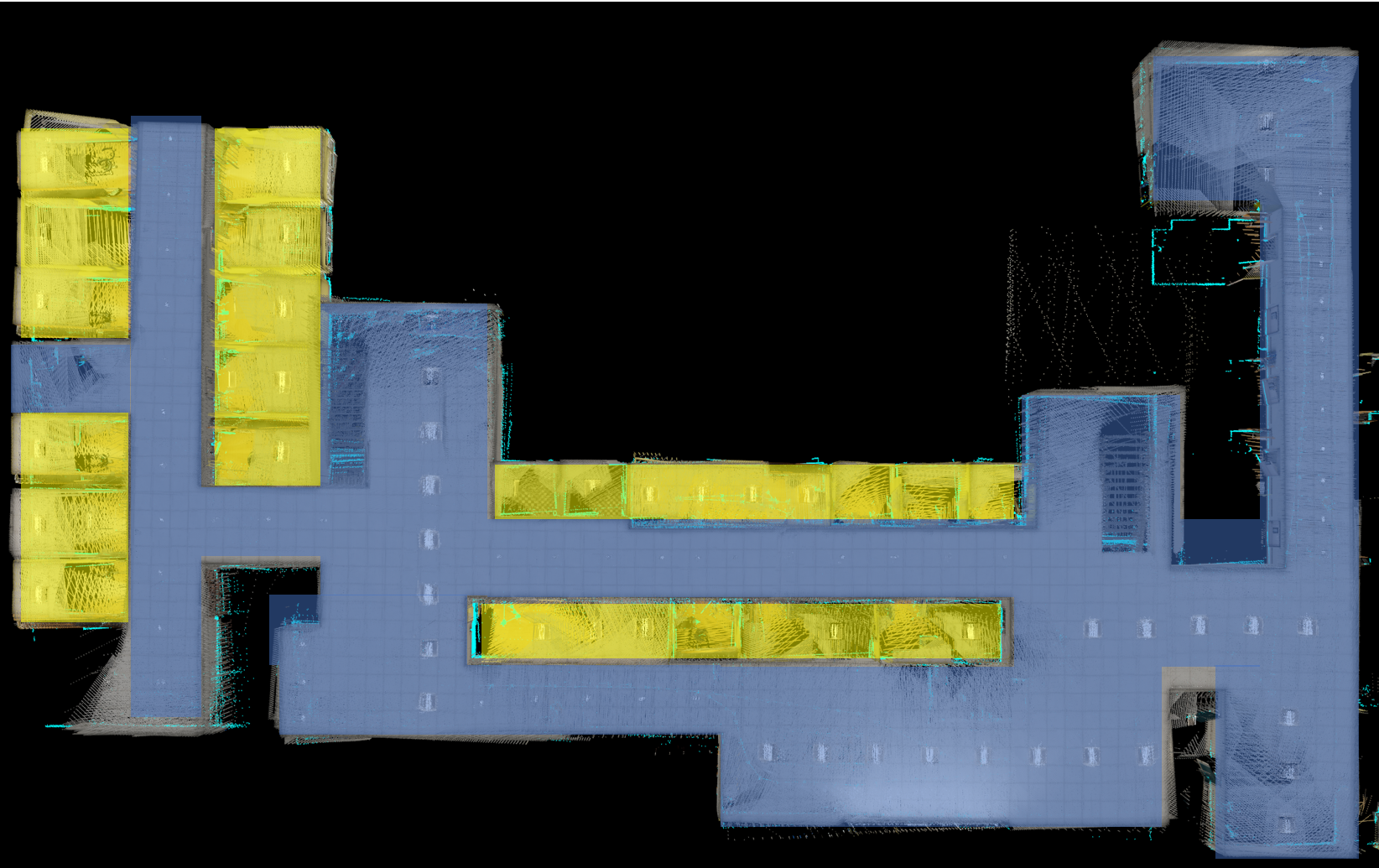}
         \caption{Amazon hospital world 3D map with layer. Blue area is Skeleton Map and yellow area is individual space layer}
         \label{figure_4_b}
     \end{subfigure}
     \hfill
     \begin{subfigure}[b]{0.40\textwidth}
         \centering
         \includegraphics[width=\linewidth]{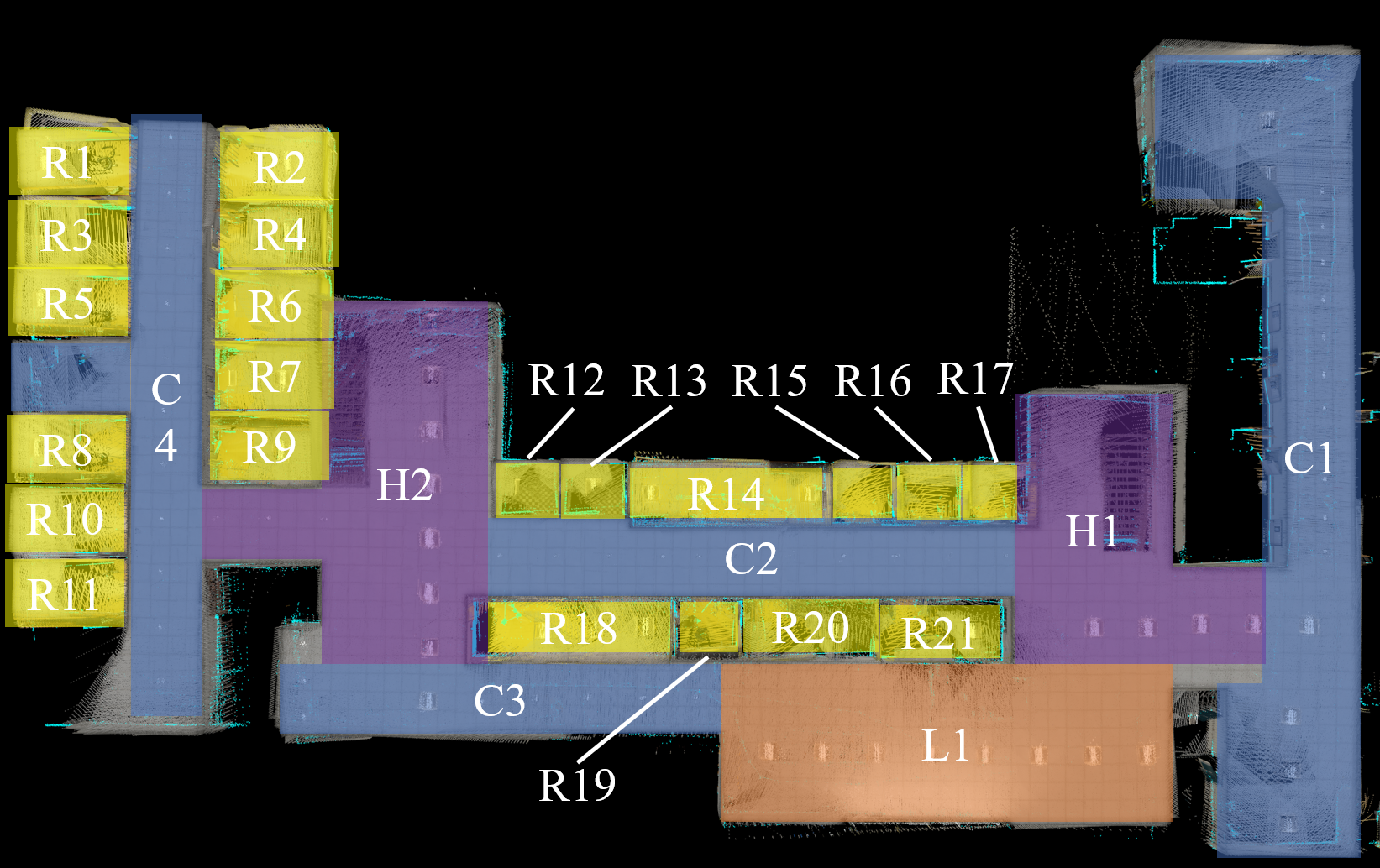}
         \caption{Amazon hospital world 3D map with semantic zone}
         \label{figure_4_c}
     \end{subfigure}
     \hfill
        \caption{3D Map with Layer, Semantic Zone}
        \label{figure_4}
\end{figure}

This research employs a hierarchical 3D mapping strategy when generating 3D maps using RTAB-Map, rather than simply scanning the entire environment at once. The specific hierarchical mapping method is as follows:

First, generate skeleton maps of the lobby and main corridor sections. Initially, mapping focuses on the lobby and main corridor areas to generate 3D base maps forming the environment skeleton. Second, add individual room layers. After skeleton maps are established, individual spaces like patient rooms and examination rooms are mapped. The maps of individual rooms are subsequently added as layers onto the existing skeleton maps.

This hierarchical mapping strategy is meaningful in two respects. First, it maintains consistency with semantic zone design. Signatures generated during the skeleton stage belong to corridor or lobby zones, while signatures generated during individual room mapping are assigned to individual room zones. As a result, zone-signature mapping is clearly separated by zone.

Second, it provides advantages for map scalability. When additional rooms or areas require mapping or modification, the existing corridor and lobby skeleton maps remain unchanged while new semantic zones and signatures for the relevant areas are edited and stacked as layers. This proves advantageous when environmental changes require map modifications or new spaces must be added. For example, building skeleton portions remain unchanged over time while being the foundation of 3D maps. Signatures from this layer are maintained unchanged as map foundations, and relatively dynamic signature layers including dynamic objects are applied on top, making hierarchical 3D maps highly scalable.

Therefore, hierarchical 3D mapping using RTAB-Map represents a mapping construction approach well-suited to the semantic zone-based memory management technique. Fig. 4(a) shows the complete 3D map, (b) shows the layers, and (c) shows the divided zones.

\subsection{Semantic Zone Based Memory Management Algorithm Implementation}

RTAB-Map's Retrieval algorithm is replaced with the semantic zone-based memory management algorithm. This algorithm follows the technique described in Section IV and is adapted to the RTAB-Map implementation environment. All other RTAB-Map components (e.g., loop closure detection) maintain existing RTAB-Map methods. Additionally, RTAB-Map's memory management operates at the signature level, with WM Size representing the count of signatures currently in WM and MemoryThr representing the maximum count of signatures to be maintained in WM. Based on this, the pseudocode of the semantic zone-based memory management algorithm is Algorithm 1.

\begin{algorithm}[t]
    \caption{Semantic Zone Based Memory Magement in RTAB-Map}
    \label{alg:zone_switching_improved}
    \begin{algorithmic}[1]
        \State \textbf{Input:} current zone $z_{\text{curr}}$, estimated pose $\hat{\mathbf{x}}_t$, active zone set $A$, WM usage $M$, WM threshold $M_{\max}$
        \State $z_{\text{new}} = \textsc{GetSwitchingZone}(\hat{\mathbf{x}}_t, z_{\text{curr}})$
        
        \\
        \If{$z_{\text{new}} \neq \texttt{null}$}
            \State $\mathcal{A} = \mathcal{A} \cup \{ z_{\text{new}} \}$
            \State $M_{pre}$ = $M$ + $|{z_{\text{new}}}|$
            \State \hspace{2em}\texttt{// Add the new zone to the active zone set  and update predict memory}

            \\
            \While{$M_{pre}$ > $M_{\max}$}
                \State $z_{\text{old}} = \textsc{SelectOldestZone}({A})$
                \State \hspace{2em}\texttt{// Select the least recently used zone in ${A}$}
                
                \State $A = {A} \setminus \{ z_{\text{old}} \}$
                \State \hspace{2em}\texttt{// Remove the oldest zone from the active zone set}
                \State $M_{pre}$ = $M_{pre}$ - $|{z_{\text{old}}}|$
                
                \State \textsc{Forget}($\mathcal{A}$)
            \EndWhile

        \EndIf
        \\
        \State \textsc{LoadZone}($z_{\text{new}}$)
    \end{algorithmic}
\end{algorithm}

If the robot's estimated position reaches a switching point between curr\_zone and new\_zone, add new\_zone to the zone set. Before loading, if the sum of the current zone set's signatures and new\_zone's signatures exceeds the set memory threshold M\_max, remove the oldest zone from the active zone set. Subsequently, perform unloading using the forget function, which transfers WM signatures to LTM. Parameters receive all zone signature IDs except the removed zone, transferring those WM signatures to LTM. Unloading of zones removed from active zone set is thereby performed, repeating until expected memory falls below M\_max. Subsequently, new\_zone is loaded.

\subsection{Test Scenarios}

Two test scenarios were designed to evaluate the map management technique based on semantic zones:
\begin{figure}[t]
    \centering
    \centerline{\includegraphics[width=80mm]{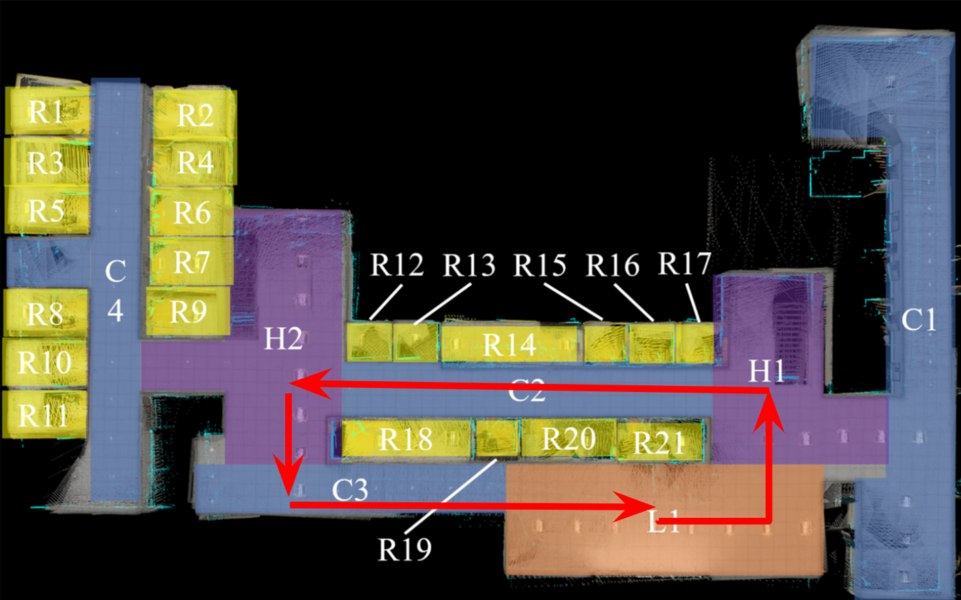}}
    \caption{Loop Scenario route}
    \label{figure_5} 
\end{figure}

\begin{figure}[t]
    \centering
    \centerline{\includegraphics[width=80mm]{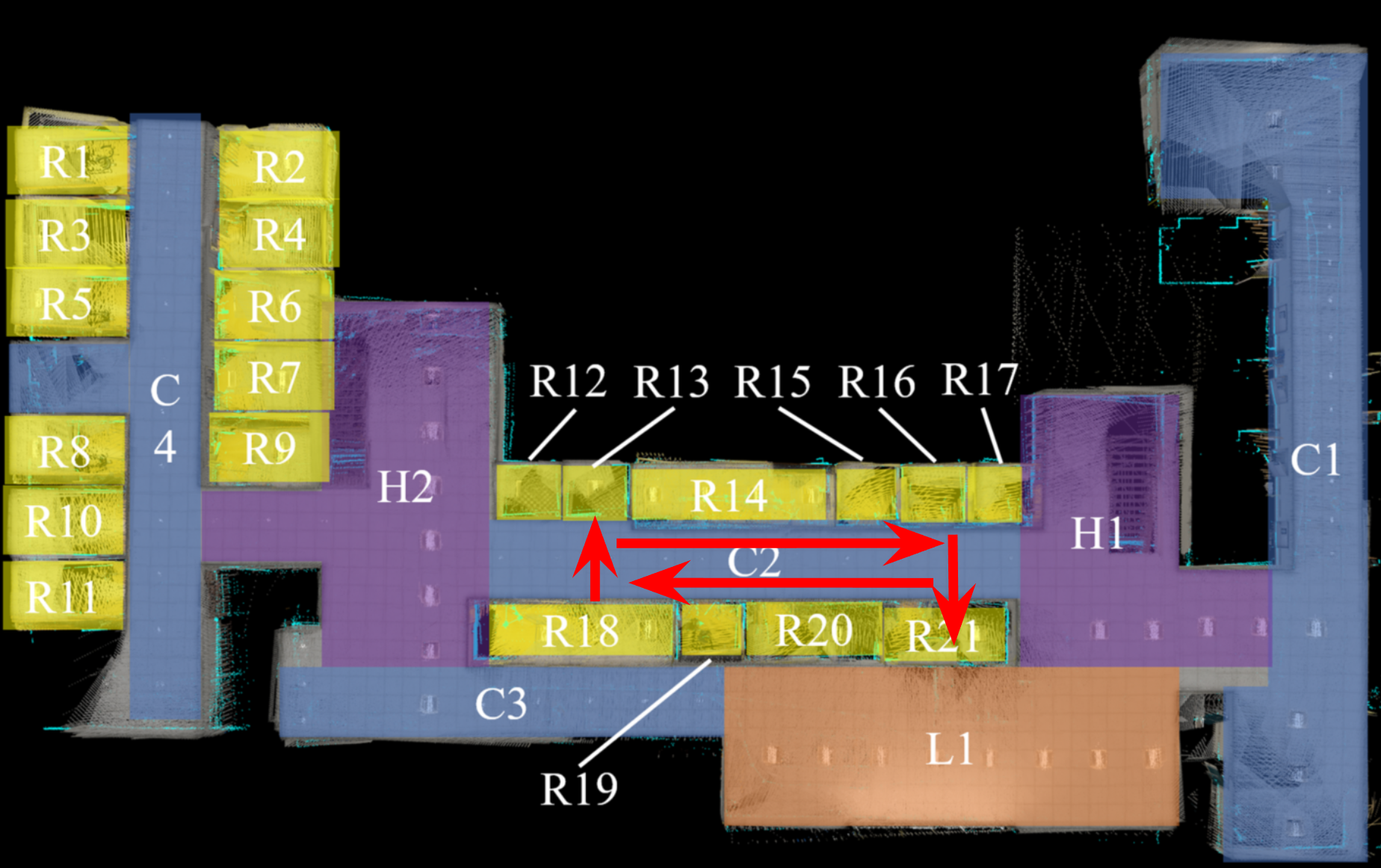}}
    \caption{Round-Trip Operation Scenario route}
    \label{figure_6} 
\end{figure}

\subsubsection{Loop Scenario}

As shown in Fig. 5, the path L1 → H1 → C2 → H2 → C3 → L1 starts from L1 and returns to L1 through H1, C2, H2, and C3. This includes relatively simple sections (L1, H1, H2, C3) and complex sections with many adjacent patient rooms (C2) where unwanted keyframes might exist. By comparing load quantities between simple structured sections and complex multi-room sections, this enables observing how the proposed technique handles memory usage patterns differently based on environmental complexity.

\subsubsection{Round-Trip Operation Scenario}

As shown in Fig. 6, the path R21 → C2 → R13 → C2 → R21 starts from R21 and returns to R21 after round-tripping through the C2-R13 section. This scenario enables zone revisits, allowing comparison of how efficiently identical signature IDs are reused between WM and LTM and how semantic zone-based management reduces unnecessary loading during this process.

\section{EVALUATION}

This section evaluates whether the proposed semantic zone-based memory management technique is actually more efficient than existing standard RTAB-Map memory management. To this end, predetermined navigation scenarios were executed in the Isaac Sim-based Amazon Hospital World environment described in Section V, and memory usage and signature load/unload patterns were compared across algorithm implementations.

\begin{figure}[t]
    \centering
    \centerline{\includegraphics[width=80mm]{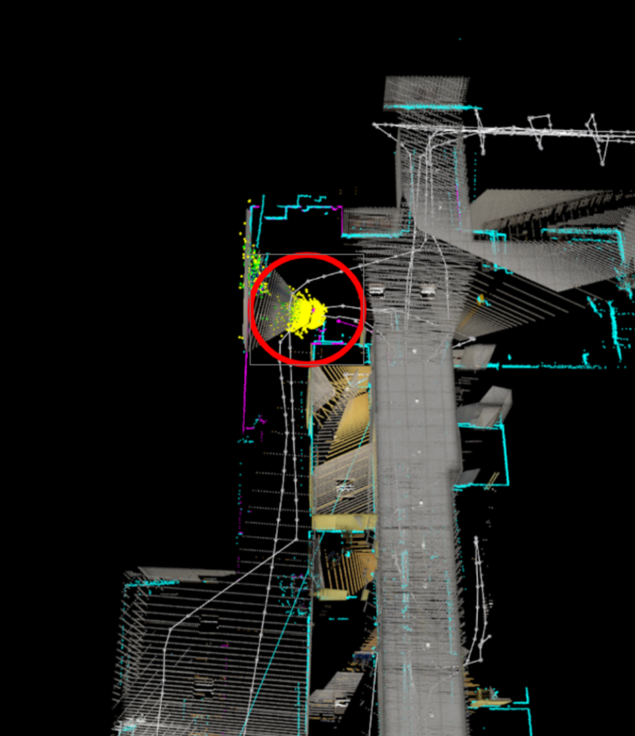}}
    \caption{Map loading slower than robot driving(1m/s)}
    \label{figure_7} 
\end{figure}

\subsection{Existing RTAB-Map Memory Management Algorithm Limitations}

\begin{figure}[!t]
     \centering
     \begin{subfigure}[b]{0.40\textwidth}
        \centering
         \includegraphics[width=\linewidth]{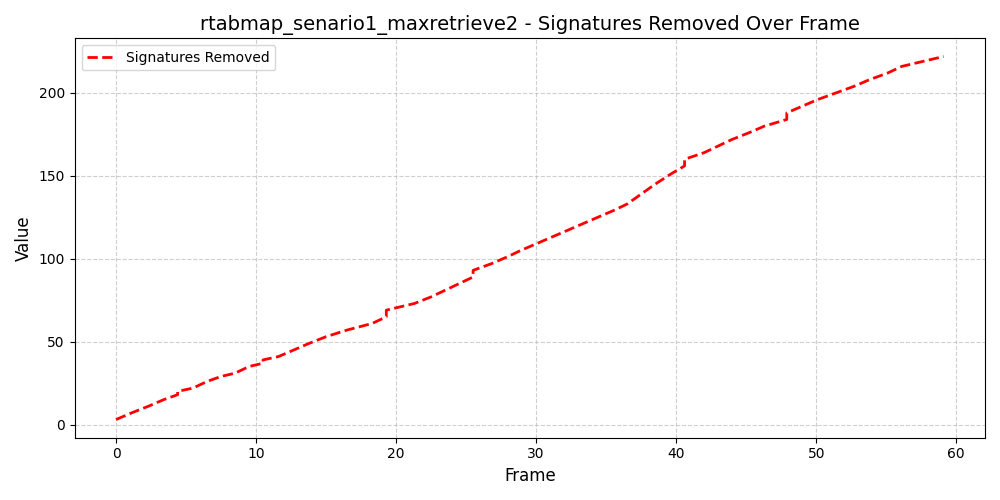}
         \caption{RTAB-Map memory management algorithm. MemoryThr 100, Maxretrieve 2 removed}
         \label{figure_8_a}
     \end{subfigure}
     \begin{subfigure}[b]{0.40\textwidth}
         \centering
         \includegraphics[width=\linewidth]{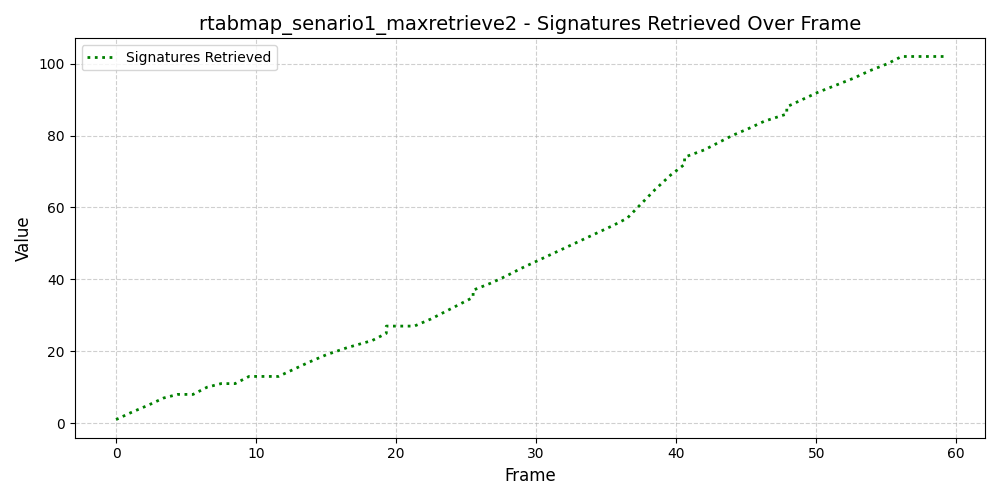}
         \caption{RTAB-Map memory management algorithm. MemoryThr 100, Maxretrieve 2 retrieve}
         \label{figure_8_b}
     \end{subfigure}
     \hfill
     \begin{subfigure}[b]{0.40\textwidth}
         \centering
         \includegraphics[width=\linewidth]{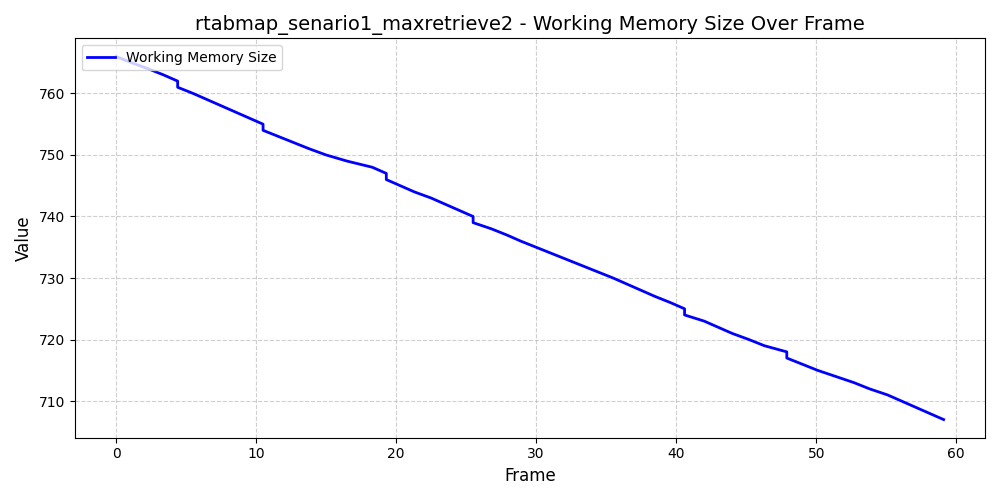}
         \caption{RTAB-Map memory management algorithm. MemoryThr 100, Maxretrieve 2 working memory size}
         \label{figure_8_c}
     \end{subfigure}
     \hfill
        \caption{RTAB-Map memory management algorithm. MemoryThr 100, Maxretrieve 2}
        \label{figure_8}
\end{figure}

RTAB-Map includes options presenting trade-offs between memory management and performance. MaxRetrieved and MemoryThr options exist, with MaxRetrieved limiting signatures retrieved per frame (RTAB-Map tick unit), preventing excessive simultaneous loading on edge devices and reducing computational burden. However, if this option is set too low, signature retrieval lags robot movement speed as in Fig. 7, resulting in slower map deployment than robot speed.

Fig. 8 shows scenario 1 performance with MaxRetrieved set to 2. The continuous linear accumulation of retrieve and remove operations indicates loop closure is not the issue; rather, insufficient retrieve quantity or loading of unnecessary signatures beyond robot traversal speed is the cause. Therefore, RTAB-Map baseline algorithm experiments were conducted with MaxRetrieved set to 10. Although retrieval speed still lags robot movement speed to some degree, increasing beyond 10 produces no performance improvement, so the parameter was fixed at 10.

RTAB-Map provides a MemoryThr parameter for specifying maximum signature counts to maintain in WM. However, this parameter does not strictly guarantee maintenance of exactly this threshold but operates as a guiding threshold triggering memory cleanup. This represents a fundamental limitation.

\subsubsection{RTAB-Map Process() Flow}

RTAB-Map fundamentally operates through the Process() function workflow, with each Process() completion representing one frame. Process() begins upon new sensor data receipt and follows this sequence:

\begin{enumerate}
\item New signatures are created and first added to WM.
\item Core SLAM computations including feature matching, pose estimation, and loop closure detection are performed. (Retrieval occurs at this point)
\item After all computations complete, WM size and processing time are verified.
\item Memory cleanup is performed based on MemoryThr, TimeThr, and other threshold exceedances.
\end{enumerate}

Due to this retrieve-first, remove-later structure, WM size structurally exceeds MemoryThr.

\begin{figure*}[t]
    \centering
    \centerline{\includegraphics[width=180mm]{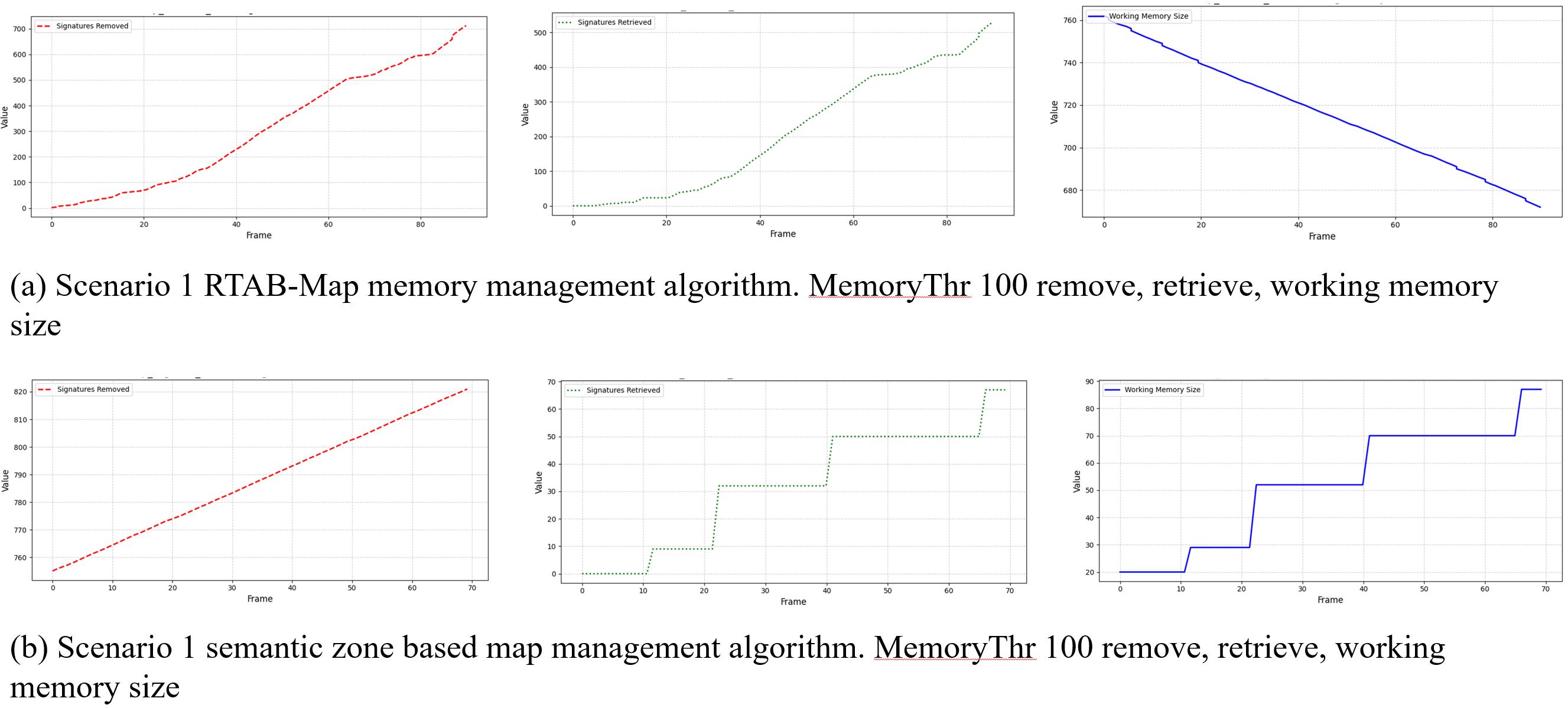}}
    \caption{Scenario 1 compare}
    \label{figure_9} 
\end{figure*}

\begin{figure*}[t]
    \centering
    \centerline{\includegraphics[width=180mm]{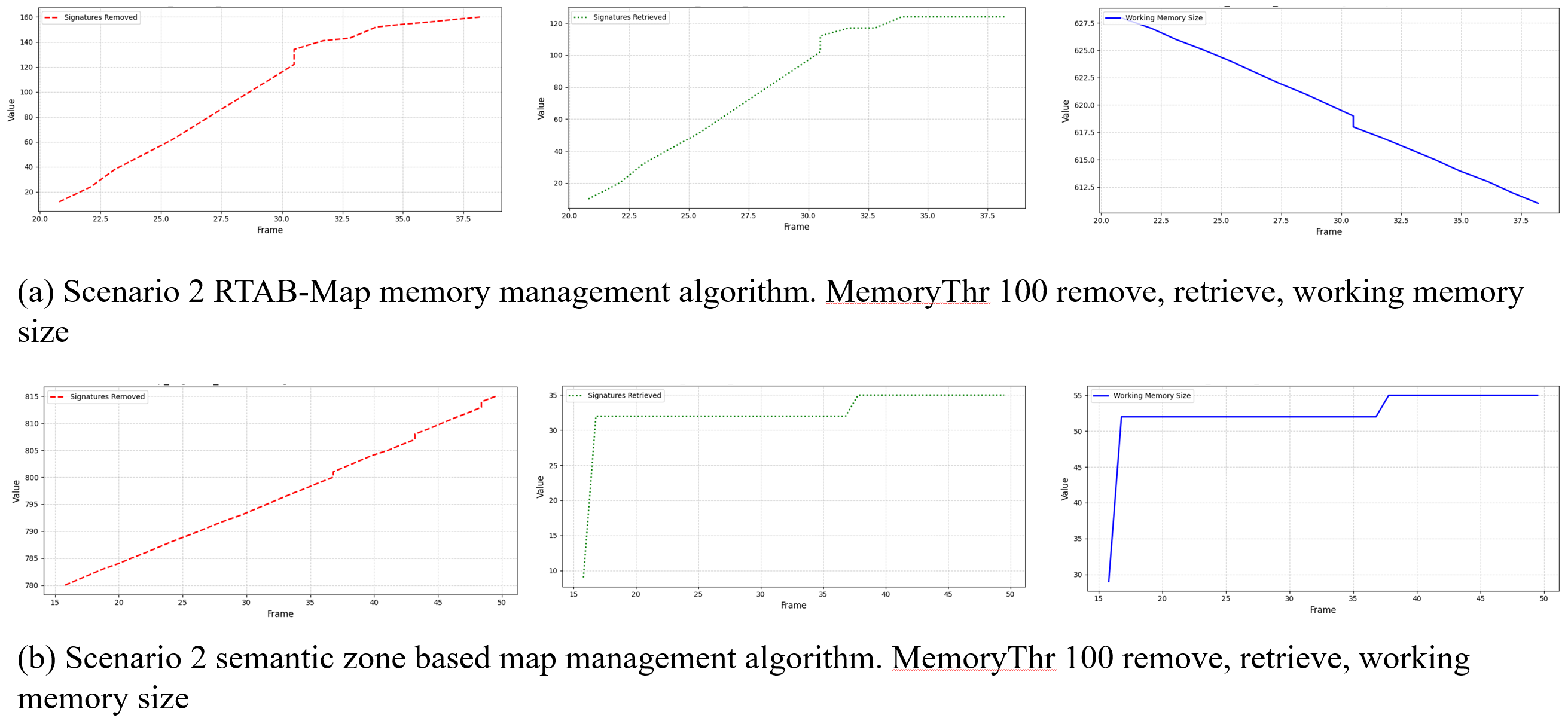}}
    \caption{Scenario 2 compare}
    \label{figure_10} 
\end{figure*}

\subsubsection{Forget() Mechanism}

RTAB-Map's memory management operates through the forget() function, which receives signatures to maintain in WM and transfers all others to Long-Term Memory (LTM). Within Process(), transfer candidate priorities are determined by time and weight criteria (nodes with high importance in loop closure or proximity detection), with fixed quantities transferred to LTM. However, this process is complicated by immunization mechanisms affecting memory management predictability.

\subsubsection{Immunize Mechanism}

\begin{figure*}[t]
    \centering
    \centerline{\includegraphics[width=180mm]{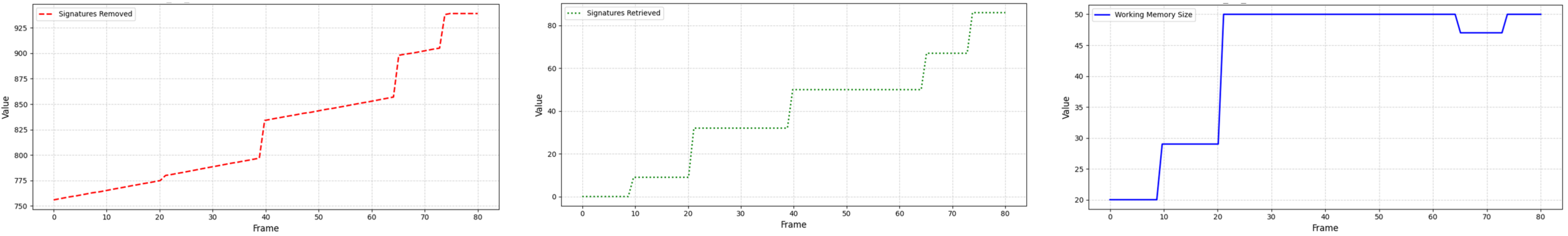}}
    \caption{Semantic zone based map management algorithm. MemoryThr 50 remove, retrieve, working memoery size}
    \label{figure_11} 
\end{figure*}

RTAB-Map applies an immunization mechanism through LocalImmunizationRatio parameter (default 0.25), immunizing nodes representing a certain percentage of current WM. Immunized nodes include those where loop closure occurred and surrounding nodes, recently added nodes still necessary for pose estimation and loop closure detection, and nodes within current robot local regions.

Immunization is essential for ensuring SLAM accuracy and stability. However, as immunized node count increases, the candidate node pool available for LTM transfer through forget() shrinks dramatically. Particularly when accumulated immunized node size approaches or exceeds MemoryThr itself, MemoryThr effectively becomes ignored.

\subsubsection{Excessive Memory Usage During Initial Localization}

During initial localization, all signatures from LTM are batch-loaded into WM. While this design rapidly determines robot position within pre-built maps, RTAB-Map's conservative transfer policy means considerable time (frames) elapses before WM size stabilizes to MemoryThr levels. Fig. 8 (c) illustrates this phenomenon.

\subsection{Signature Load/Unload Cumulative Count Comparison}

Comparison of signature load/unload cumulative counts between baseline RTAB-Map and semantic zone-based 3D Map Management demonstrates the efficiency of signature load/unload usage in semantic zone approaches. Parameters used were MaxRetrieve 10 and MemoryThr 100. Additionally, baseline RTAB-Map scenario execution aligned signature retrieve speed with robot movement speed, while semantic zone-based approaches included slight delays in estimated position coordinate calculations and corresponding zone loading, executing scenarios while awaiting these delays.

\subsubsection{Scenario 1 - Loop Scenario Comparison}

This compares accumulated load/unload quantities during loop scenario 1. During baseline RTAB-Map algorithm testing in scenario 1, the C2 region (complex multi-room area) was traversed between frames 30-60. Fig. 9(a) shows considerable loading during this interval. Indeed, due to signature density in complex regions, loading and resulting unloading from loading increases significantly. The baseline RTAB-Map algorithm ultimately performed approximately 700 unload and 520 load operations during scenario 1. 

In contrast, the semantic zone approach pre-maps zone-signature associations, loading only predetermined zone signatures. Once zones are loaded, no additional loading occurs until reaching the next zone. Since signature totals for scenario 1 zones did not exceed 100, no memory-exceeding removal occurred. Only frame generation and removal each frame during localization operation remained. The semantic zone-based memory management technique ultimately performed approximately 70 unload and 80 load operations during scenario 1.

\subsubsection{Scenario 2 - Round-Trip Operation Scenario Comparison}

Differences are more pronounced in scenario 2 round-trip operation. Fig. 10(b) shows that semantic zone approach retrieval is observed only for three keyframes required to load R13, with no other retrieve operations observable. However, Fig. 10(a) shows baseline RTAB-Map algorithm continuously performing load/unload operations. This demonstrates that signature load/unload quantities substantially decrease in round-trip operation scenarios.

\subsection{Semantic Zone with Smaller MemoryThr}

Matching scenario execution methods from Section 6.2 but with reduced MemoryThr values, the semantic zone-based memory management technique was tested with MemoryThr set to 50 for scenario 1 performance.

Fig. 11 shows proper remove algorithm operation, while confirms memory threshold compliance.

\section{DISCUSSION}

This research introduced semantic zone-based memory management techniques into RTAB-Map's existing memory management structure, achieving memory threshold guarantees previously absent in baseline RTAB-Map while achieving dramatic performance improvements in accumulated signature load/unload quantities. Particularly in hospitals with multiple rooms adjacent to corridors, this research addresses the inefficiency of frequently loading signatures from never-entered rooms, with semantic zone-based approaches structurally reducing such inefficiencies.

However, this research has limitations in using manually defined semantic zones. Since pre-defined zones are used, application is restricted to localization modes using pre-built maps rather than real-time mapping modes, and manually defined zones may not represent optimal partitioning. However, based on Zone-based Federated Learning research, zone improvements appear possible by dynamically adjusting zone sizes and boundaries according to actual robot movement patterns and usage frequency.

Another constraint of semantic zone-based memory management involves global path planning for currently inactive zones. Zone-based approaches reduce memory through selective loading, potentially complicating path planning toward distant targets.

This research applied zone-level loading and MemoryThr-based unloading techniques in experimental comparisons. Further research into zone-level removal techniques appears necessary. Applying existing computer memory management algorithms or movement prediction-based memory management upon accumulated robot movement data could further enhance performance.

\section{CONCLUSION}

This thesis proposes and validates a semantic zone-based 3D map management technique addressing memory constraint problems encountered when mobile robots operate 3D maps in large-scale indoor environments such as hospitals and warehouses. While existing SLAM frameworks primarily depend on temporal ordering or geometric distance for map data management, suffering from frequent unnecessary loading/unloading in complex indoor environments, this research overcomes these limitations by partitioning environments into functionally meaningful semantic zones including lobbies, corridors, and patient rooms, selectively maintaining only necessary zones in memory based on robot location, thereby solving memory management challenges.

The primary conclusions and achievements of this research are as follows. First, introducing the semantic zone concept improved memory management efficiency. By managing memory based on spatial semantic boundaries, unnecessary loading of adjacent space data never entered by robots was structurally prevented. This demonstrated stable 3D map utilization for large-scale environments even within mobile robots with limited resources. Second, implementation and experimentation on the universal SLAM framework RTAB-Map verified actual operational feasibility. Integration with RTAB-Map's WM and LTM structures and experimentation in virtual hospital environments based on Isaac Sim demonstrated practical viability.

Future research directions include automating zone setting currently requiring manual user zone-keyframe mapping through semantic mapping integration, and when deployed on actual mobile robots, dynamically improving zones by reflecting actual data in zone sizes and boundaries. Additionally, while current focus lies on zone-level loading/unloading and strict memory management, future work should extend toward prediction-based memory management algorithms that predict robot movement paths to proactively remove unnecessary zones or preload anticipated zones.

In conclusion, this research proposes semantic zone-based 3D map management as a solution for enabling mobile robots to efficiently utilize 3D maps in large-scale complex indoor spaces. This is anticipated to contribute to enabling mobile robots to overcome hardware constraints and enhance technical completeness in hospitals, warehouses, and other large-scale environments.

\addtolength{\textheight}{-12cm}

\end{document}